\documentclass[sigconf, nonacm]{acmart}

\usepackage{booktabs}
\usepackage{multirow}
\usepackage{xcolor}
\usepackage{algorithm}
\usepackage{algorithmic}

\newcommand{\sys}{DeFT-RAG}
\newcommand{\lsub}{\ensuremath{\mathrm{LLM}_{\mathrm{sub}}}}
\newcommand{\lans}{\ensuremath{\mathrm{LLM}_{\mathrm{ans}}}}
\newcommand{\lfin}{\ensuremath{\mathrm{LLM}_{\mathrm{fin}}}}

\begin{document}

\title{DeFT-RAG: Decomposing Multi-Hop Questions into Self-Contained Sub-Questions with Reusable Answers}

\author{Dingyi Kang}
\affiliation{%
  \institution{University of Texas at Dallas}
  \city{Richardson}
  \state{Texas}
  \country{USA}}
\email{dingyi.kang@utdallas.edu}

\author{Bingzhe Li}
\affiliation{%
  \institution{University of Texas at Dallas}
  \city{Richardson}
  \state{Texas}
  \country{USA}}
\email{bingzhe.li@utdallas.edu}

\begin{abstract}
Retrieval-augmented generation (RAG) lets a large language model (LLM) answer questions by retrieving relevant passages from a large corpus. However, RAG struggles with multi-hop questions, which require chaining facts spread across several documents through intermediate entities that the question never mentions. Graph-based RAG methods retrieve once from the entities a question names and so rarely recover the full evidence chain. Iterative RAG methods recover it step by step, but their accuracy is limited by the many tasks they ask the LLM to perform, such as planning the sub-questions, deciding their dependencies, rewriting them as answers arrive, and keeping track of what is known. We propose DeFT-RAG, a RAG framework built on three components. First, DeFT-RAG generates one self-contained sub-question at a time from the original question and the sub-questions answered so far with their answers (QA pairs). Each sub-question names every entity it asks about, which removes the need to plan, track dependencies or rewrite, and lets the LLM focus on a single fine-grained task at a time. Second, DeFT-RAG moves the remaining LLM work outside the LLM: rule-based updates to a QA table that keeps track of what is known, and yes/no decisions made by a fine-tuned lightweight decision model. Third, DeFT-RAG caches the QA pairs and reuses an answer when a later question asks a sub-question with the same meaning. On MuSiQue, 2WikiMultiHopQA and HotpotQA, DeFT-RAG achieves the best accuracy among thirteen baselines with both gpt-4o-mini and Qwen3.6 as the backbone, while giving the LLM fewer input tokens than any other iterative method. With a QA cache shared across 1{,}000 MuSiQue questions, 23.1\% of sub-questions are answered from the cache, which reduces LLM calls by 12.9\% without lowering accuracy.
\end{abstract}



\maketitle

\section{Introduction}
\label{sec:intro}
\begin{figure*}[t]
\centering
\includegraphics[width=\textwidth]{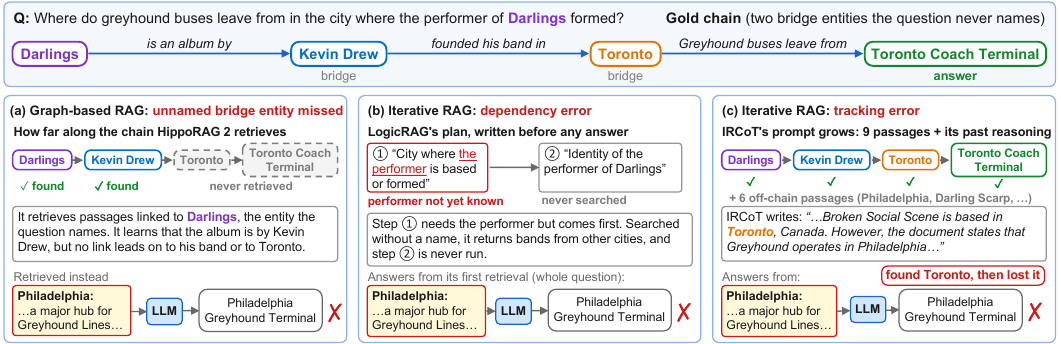}
\caption{Failures of graph-based and iterative RAG on a 3-hop MuSiQue question (gpt-4o-mini runs). Top: the gold chain, which passes through two bridge entities the question never names. (a) HippoRAG~2, a graph-based method, finds the first bridge entity (Kevin Drew) but no passage about his band or Toronto. (b) LogicRAG, which plans its sub-questions in advance, searches for the city of ``the performer'' before finding the performer (dependency error). (c) IRCoT, which keeps all retrieved passages in its long prompt, has the whole chain there and even states that the band is based in Toronto, but then loses track of this fact (tracking error).}
\label{fig:motivation}
\end{figure*}

Large language models (LLMs) now perform strongly across a wide range of language tasks~\cite{openai2023gpt4,yang2025qwen3}. Yet because their knowledge is stored in their parameters, they often answer incorrectly when that knowledge is outdated or missing, and they may hallucinate~\cite{huang2025hallucination}. Retrieval-augmented generation (RAG)~\cite{lewis2020rag,gao2023ragsurvey} mitigates this by retrieving passages from an external corpus and grounding the answer in them, which improves accuracy on knowledge-intensive tasks without retraining the model.

A single retrieve-then-generate step suffices when one passage holds the answer, but it falls short on multi-hop question answering~\cite{yang2018hotpotqa,ho2020constructing,trivedi2022musique}, where answering one question requires chaining several facts spread across different documents: retrieving with the question alone returns passages that match its surface terms, and these rarely cover every link of the chain. Graph-based RAG methods~\cite{edge2024graphrag,gutierrez2024hipporag,gutierrez2025hipporag2,guo2024lightrag,chen2025pathrag,wu2026memgraphrag} address this by building a graph over the corpus that links entities and passages, then retrieving passages by following links from the entities in the question. However, a multi-hop chain runs through \emph{bridge entities} that the question never names and that only become known once an earlier step is answered. Because most graph-based methods retrieve once per question, starting from the entities the question names, they can reach the bridge entities only by following graph links from those entities, which rarely gets past the first one: for most of these methods, the full evidence chain is retrieved for only 6--10\% of 3-hop MuSiQue questions (Figure~\ref{fig:motivation}a, Table~\ref{tab:coverage-full}).

Iterative RAG methods~\cite{trivedi2023ircot,press2023selfask,chen2026logicrag,kim2026hiq} instead aim to find these bridge entities directly: they decompose the question into sub-questions and alternate between retrieval and LLM reasoning, so each answered sub-question supplies the bridge entity needed by the next retrieval. However, their accuracy is limited by the many tasks they ask the LLM to perform, often within one call, such as planning the sub-questions, deciding the dependencies among them, rewriting them as answers arrive, and keeping track of what is known, each of which leads to its own type of error. Specifically, a plan written before any sub-question is answered must guess bridge entities, and a wrong guess or a missing step misleads every later step (\emph{planning errors}). The dependencies decided among the sub-questions may link them wrongly or in the wrong order, so a sub-question is searched before the fact it needs is found (\emph{dependency errors}), as in Figure~\ref{fig:motivation}b. When an answer arrives, the LLM may fail to rewrite the sub-question that depends on it, which keeps a placeholder instead of the entity's name, or may rewrite it with the wrong entity (\emph{rewriting errors}). In a long, growing input, the LLM may also lose track of what it has already found (\emph{tracking errors}), as in Figure~\ref{fig:motivation}c. Moreover, fixed top-$k$ retrieval, used by most RAG methods, makes these issues worse by lowering the answer accuracy at each step. Finally, iterative methods discard their intermediate results once a question is answered, although multi-hop questions over the same corpus often share intermediate facts that could be reused.

Based on these observations, we propose \sys{}, a RAG framework built on three components. First, \sys{} generates one self-contained sub-question at a time from the original question and the sub-questions answered so far with their answers (QA pairs), until these accumulated QA pairs are enough to answer the original question. Each sub-question names every entity it asks about and does not depend on the answer to any other sub-question. Since nothing is planned in advance, no bridge entity has to be guessed before it is found, and there are no dependencies to get wrong and no sub-question to rewrite. This avoids planning, dependency and rewriting errors, and lets the LLM focus on a single fine-grained task at a time: writing a sub-question for the next missing piece of information. Second, \sys{} converts the remaining LLM work into fine-grained tasks handled outside the LLM. A QA table keeps every QA pair and is updated by rules, without any model call, so the LLM no longer has to keep track of what is known, which avoids tracking errors. A fine-tuned lightweight decision model (DM), a language model trained to answer yes/no questions with a probability~\cite{startlux2026}, makes a few yes/no decisions: whether a question can be answered directly, whether the QA pairs found so far are enough, and whether retrieved passages are relevant, so that the LLM sees more of the needed passages and fewer distracting ones. Third, \sys{} caches the QA pairs and reuses an answer when a later question asks a sub-question with the same meaning, which saves retrieval and LLM calls. Together, these designs raise the accuracy of each step and of the final answer, while making the intermediate answers reusable across questions. Our contributions are summarized as follows:
\begin{itemize}
\item We propose an incremental question decomposition that generates one self-contained sub-question at a time from the original question and the QA pairs found so far. It removes the need to plan sub-questions, decide their dependencies or rewrite them, and lets the LLM focus on a single fine-grained task: writing the sub-question for the next missing piece of information.
\item We convert the remaining LLM work into fine-grained tasks handled outside the LLM: rule-based updates to a QA table, and yes/no decisions by a fine-tuned lightweight DM, such as whether retrieved passages are relevant. This aims to raise the accuracy of each step, and to leave the LLM only a few focused, fine-grained tasks that require text generation.
\item We design a QA cache that reuses the answers to earlier sub-questions for later questions that ask a sub-question with the same meaning, which saves retrieval and LLM calls.
\item We evaluate \sys{} on three multi-hop QA benchmarks with two LLM backbones. It achieves the best accuracy among thirteen baselines while giving the LLM fewer input tokens than any other iterative method, and reusing answers through the QA cache further reduces LLM calls without lowering accuracy. To support reproducibility, we release our code and prompts at \href{https://github.com/Dingyi-Kang/DeFT-RAG}{\nolinkurl{https://github.com/}\linebreak[0]\mbox{\nolinkurl{Dingyi-Kang/DeFT-RAG}}}.
\end{itemize}

\section{Background and Problem Formulation}
\label{sec:background}

This section introduces RAG and multi-hop question answering, and the limitations of existing methods on such questions; Section~\ref{sec:related} discusses related work in more detail.

\subsection{Retrieval-Augmented Generation}
\label{sec:bg-rag}
Given a corpus $\mathcal{C}$ of passages and a question $Q$, a vanilla RAG system answers
\begin{equation}
  a = \mathrm{LLM}\big(Q,\, R(Q, \mathcal{C})\big),
  \label{eq:rag}
\end{equation}
where $R$ retrieves the passages relevant to $Q$ from $\mathcal{C}$. Vanilla RAG implements $R$ as similarity search, whether by dense-embedding cosine or sparse lexical matching (BM25), ranking passages by their similarity to the query:
\begin{equation}
  R_k(q, \mathcal{C}) = \operatorname*{top\text{-}k}_{p \in \mathcal{C}}\; s\big(e(q),\, e(p)\big),
  \label{eq:topk}
\end{equation}
where $e(\cdot)$ maps a query or passage to its representation and $s$ is a similarity function. We refer to Eq.~\eqref{eq:topk} as \emph{fixed top-$k$ similarity retrieval}: the number of passages $k$ is fixed in advance, and passages are ranked only by their similarity to the query.

\subsection{Multi-Hop Question Answering}
\label{sec:bg-mhqa}
A multi-hop question usually cannot be answered from a single passage~\cite{yang2018hotpotqa,ho2020constructing,trivedi2022musique}. Answering it requires an \emph{evidence chain} of two or more facts, each stated in a passage of $\mathcal{C}$, where some facts can only be found once an earlier one is known. We call the intermediate answers that later facts depend on \emph{bridge entities}: the chain passes through them, but $Q$ never names them. For example, to answer ``Where do greyhound buses leave from in the city where the performer of Darlings formed?'', one must first find that Darlings is an album by Kevin Drew, then that he formed his band, Broken Social Scene, in Toronto, and only then where Greyhound buses leave from there (the Toronto Coach Terminal). Kevin Drew and Toronto are the bridge entities (top of Figure~\ref{fig:motivation}).

Vanilla RAG (Eq.~\eqref{eq:rag}) handles such questions poorly. Eq.~\eqref{eq:topk} ranks passages only by their similarity to $Q$, so for the Darlings question it returns passages that share its words, such as passages about greyhound dogs, a Greyhound coach model and a person named Darling. Answering $Q$, however, also needs passages about Kevin Drew and Toronto. Because $Q$ never mentions either of them, their passages rank low, and a single retrieval rarely returns the full evidence chain: on MuSiQue, vanilla RAG retrieves it for only 7\% of 3-hop questions (Table~\ref{tab:coverage-full}). To address this, two lines of work have been developed: graph-based RAG (\S\ref{sec:bg-graph}) and iterative RAG (\S\ref{sec:bg-iter}).

\subsection{Graph-Based RAG}
\label{sec:bg-graph}
Graph-based RAG methods build a graph over $\mathcal{C}$ offline, either by using an LLM to extract the entities in every passage and the relations between them~\cite{edge2024graphrag,gutierrez2024hipporag}, or by linking passages across documents through the entities they share~\cite{zhuang2025linearrag}. At answering time, they locate the entities mentioned in $Q$ in this graph and follow its links to collect related passages, instead of ranking passages only by their similarity to $Q$ as in Eq.~\eqref{eq:topk}. In this way, evidence linked to the question's entities can be retrieved even when it shares little wording with $Q$.

However, graph-based RAG still falls short in covering the full evidence chain once a question needs more than two hops: on MuSiQue, most graph-based methods retrieve the full chain for only 6--10\% of 3-hop questions, barely above vanilla RAG (Table~\ref{tab:coverage-full}); LinearRAG reaches 28\%, but only by giving the LLM about 4{,}100 words of passages per question, nearly five times as many as vanilla RAG, most of which are not part of the evidence chain. The reason is that most graph-based methods retrieve only once per question, starting from the entities $Q$ names, while the bridge entities are by definition not mentioned in $Q$, so the retrieval can reach them only by following graph links from those entities, which rarely gets past the first bridge entity. In Figure~\ref{fig:motivation}a, HippoRAG~2 learns from the album passage that Darlings is by Kevin Drew, but retrieves no passage about his band or Toronto, and answers from a passage about Greyhound in Philadelphia.

\begin{figure*}[t]
\centering
\includegraphics[width=\textwidth]{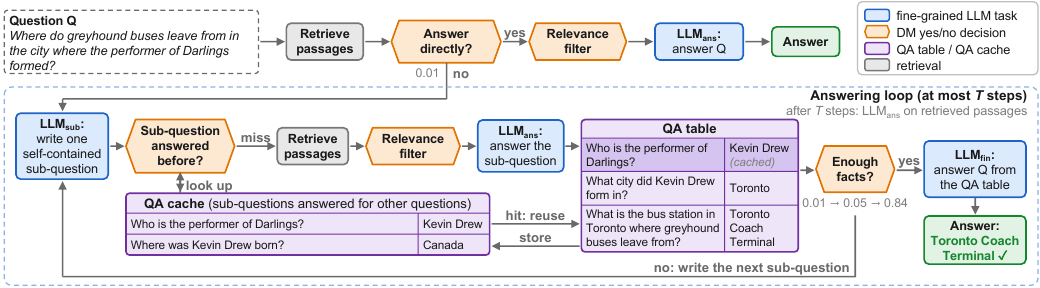}
\caption{Overview of \sys{} on the question of Figure~\ref{fig:motivation}. The LLM performs only three fine-grained tasks (blue), and a fine-tuned DM makes every yes/no decision (orange). The QA table, updated by rules, holds the QA pairs found for the question; the QA cache holds those found for other questions. The first sub-question was answered for an earlier question, so its answer is reused from the QA cache (marked ``cached'') without retrieval or an LLM call. The values next to \emph{Answer directly?} and \emph{Enough facts?} are the DM's yes probabilities from our gpt-4o-mini run.}
\label{fig:overview}
\end{figure*}

\subsection{Iterative RAG}
\label{sec:bg-iter}
Iterative RAG methods instead aim to find these bridge entities directly: rather than preprocessing the corpus, they decompose $Q$ into sub-questions and alternate between retrieval and LLM reasoning, so that each sub-question can name the bridge entity found in an earlier step and reach passages that a search with $Q$ alone would miss. As shown in Table~\ref{tab:coverage-full}, iterative methods retrieve the full evidence chain for 19--41\% of 3-hop questions, well above most graph-based methods.

However, iterative methods often fail, because they ask the LLM to perform several tasks, often within one call: planning sub-questions before any of them is answered, deciding the dependency links among them, rewriting them as answers arrive, and keeping track of what is known. Each of these tasks leads to its own type of error. First, \emph{planning errors}: methods that plan the sub-questions in advance, such as LogicRAG~\cite{chen2026logicrag} and REAP~\cite{zhu2026reap}, must guess bridge entities before finding them, and a wrong guess or a missing step misleads every later step. For example, asked ``Who is the owner of the record label of the guitarist who performed on the Top Gun theme?'', LogicRAG's plan asks for the label of the theme's composer instead of first finding the guitarist, so it answers with the owner of the wrong label. Second, \emph{dependency errors}: methods that link the sub-questions by their dependencies, such as LogicRAG and REAP, may link them wrongly or in the wrong order, so a sub-question is searched before the fact it depends on is found. In Figure~\ref{fig:motivation}b, LogicRAG searches for the city where ``the performer'' formed before it has found the performer, retrieves passages about bands from other cities, and never searches for the performer at all.

Third, \emph{rewriting errors}: when the answer to a sub-question arrives, methods such as Hi-Q~\cite{kim2026hiq} and REAP must rewrite the sub-question that depends on it. The LLM may fail to do so, and the sub-question keeps a placeholder instead of the entity's name, so retrieval often cannot find the passages about this bridge entity; or it may rewrite the sub-question with the wrong entity. Fourth, \emph{tracking errors}: methods that carry their whole reasoning history in the prompt, such as IRCoT~\cite{trivedi2023ircot} and Self-Ask~\cite{press2023selfask}, must find what is already known within a long and growing input, and may miss a fact they have already found or ask for it again. In Figure~\ref{fig:motivation}c, IRCoT's input holds all three gold passages, and its reasoning states that the band is based in Toronto, yet it then follows a passage about Greyhound in Philadelphia and answers with the Philadelphia terminal.

Moreover, fixed top-$k$ retrieval (Eq.~\eqref{eq:topk}), used by most RAG methods, makes these issues worse, because it lowers the answer accuracy of every step. With a small $k$, part of the needed evidence is missed; with a large $k$, the evidence is buried among distracting passages. Table~\ref{tab:window} illustrates this on MuSiQue questions that can be answered in one retrieval step: widening the list from the top 5 to the top 10 passages raises how often all needed passages are present from 62\% to 88\%, yet accuracy barely changes (0.620 to 0.630); keeping only the gold passages from the top 10 raises it to 0.793. 2Wiki shows the same pattern. Since iterative methods answer several sub-questions in sequence, and each step depends on the answers before it, these errors add up across steps and greatly increase the chance that the whole process fails.

\begin{table}[t]
\centering
\small
\caption{Accuracy on questions that can be answered in one retrieval step, when the LLM is given the top-$k$ retrieved passages vs.\ only the gold passages among them (sampled questions: MuSiQue $n{=}92$, 2Wiki $n{=}135$; gpt-4o-mini reader). Acc.\ = LLM-Acc; Gold = share of questions whose gold passages are all among the passages given to the LLM.}
\label{tab:window}
\begin{tabular}{lcccc}
\toprule
 & \multicolumn{2}{c}{MuSiQue} & \multicolumn{2}{c}{2Wiki} \\
\cmidrule(lr){2-3}\cmidrule(lr){4-5}
Passages & Acc. & Gold & Acc. & Gold \\
\midrule
Top-5 & 0.620 & 62\% & 0.733 & 87\% \\
Top-10 & 0.630 & 88\% & 0.748 & 98\% \\
Top-5 + gold only & 0.663 & 62\% & 0.800 & 87\% \\
Top-10 + gold only & 0.793 & 88\% & 0.874 & 98\% \\
\bottomrule
\end{tabular}
\end{table}

Finally, compared with vanilla RAG and graph-based RAG, iterative methods spend many more retrieval and LLM calls to answer a question (Appendix~\ref{app:overhead}), which are expensive in latency and cost, yet once a question is answered, they discard the intermediate sub-questions and answers. This is wasteful: multi-hop questions over the same corpus often share intermediate facts, so these answers could be reused to save retrieval and LLM calls for later questions.

\section{\sys{} Design}
\label{sec:design}
\sys{} is built on three components (Figure~\ref{fig:overview}): incremental question decomposition (\S\ref{sec:design-units}); fine-grained tasks outside the LLM, namely rule-based updates to a QA table (\S\ref{sec:design-facts}) and yes/no decisions made by a fine-tuned lightweight decision model (DM) (\S\ref{sec:design-decisions}); and reusing answers across questions through a QA cache (\S\ref{sec:design-reuse}). \S\ref{sec:design-process} shows how they work together to answer a question. We call a task fine-grained when it does only one thing at a time, from a short input, such as making one yes/no decision, or generating a new sub-question without the need to rewrite the others.

\subsection{Incremental Question Decomposition}
\label{sec:design-units}
In \sys{}, the LLM performs only three fine-grained tasks:
\begin{equation}
  q_{t+1} = \lsub(Q, F_t),\;\; a = \lans(q, P),\;\; \hat{a} = \lfin(Q, F)
\label{eq:units}
\end{equation}
where $F_t = \{(q_1, a_1), \dots, (q_t, a_t)\}$ is the QA table, the set of QA pairs (sub-questions with their answers) found after $t$ steps, and $P$ is the set of passages selected for a question (\S\ref{sec:design-decisions}). \lsub{} writes the next sub-question, \lans{} answers a question from its selected passages, and \lfin{} answers the original question from the QA table (blue in Figure~\ref{fig:overview}).

\lsub{} generates one self-contained sub-question at a time from the original question and the QA pairs found so far, asking for the next missing piece of information, until the DM judges these QA pairs enough to answer the original question (\S\ref{sec:design-decisions}). A sub-question is \emph{self-contained} when it names every entity it asks about, by name or by a description that needs no other lookup, and does not depend on the answer to any other sub-question, so it can be searched and answered on its own. For example, once the QA table shows that the Rhine ends in the North Sea, ``Where does the North Sea meet the Atlantic?'' is self-contained, while ``Where does that sea meet the Atlantic?'' is not, because ``that sea'' points to an earlier step. Since nothing is planned in advance, no bridge entity has to be guessed before it is found, which avoids the planning errors of \S\ref{sec:bg-iter}. Since no sub-question depends on another, there are no dependencies to get wrong and no sub-question to rewrite, which avoids dependency and rewriting errors. More importantly, the LLM can focus on a single fine-grained task at a time: writing a sub-question for the next missing piece of information. Together, these improve the quality of the question decomposition and help \sys{} find every bridge entity of the evidence chain: it retrieves the full chain for more questions with three or more hops than any other system in Table~\ref{tab:coverage-full}.

\subsection{QA Table}
\label{sec:design-facts}
\sys{} converts the rest of the work that iterative methods leave to the LLM into fine-grained tasks handled outside the LLM: rule-based updates to a QA table, described here, and yes/no decisions made by the DM (\S\ref{sec:design-decisions}). This aims to raise the accuracy of each step, and to leave the LLM only the three tasks in Eq.~\ref{eq:units}, which require text generation.

Instead of tracking everything found so far within the LLM's context, in a long and growing prompt of passages and past reasoning, \sys{} keeps it in a compact form, the QA table $F_t$ of Eq.~\ref{eq:units}, which starts empty, $F_0 = \emptyset$, and grows by one QA pair at each step that finds an answer:
\begin{equation}
  F_{t+1} = F_t \cup \{(q_{t+1}, a_{t+1})\}.
  \label{eq:facts}
\end{equation}
The QA table is updated by rules, without any LLM or DM call: the QA pair of each answered sub-question is added to the table; a sub-question that gets no answer is recorded, and \lsub{} is shown it so that it is not asked again; and a new sub-question that repeats one already asked is detected by comparing the normalized wording, and \lsub{} is asked for another one. Figure~\ref{fig:overview} shows the QA table for an example question. Every answer found is thus kept explicitly as a QA pair, rather than buried in a long prompt, which avoids tracking errors. The QA table is the only record of what is known passed from one step to the next: \lsub{} reads it to write the next sub-question, the DM reads it to decide whether enough is known to stop, and \lfin{} reads it to answer the original question. Since each QA pair is a self-contained sub-question with a short answer, these inputs stay short and name every entity found so far, instead of carrying long passages or the wording of earlier steps.

\subsection{Decision Tasks}
\label{sec:design-decisions}
The DM is a language model, usually with far fewer parameters than the backbone LLM, trained to answer yes/no questions with a probability~\cite{startlux2026} and fine-tuned for our decisions (\S\ref{sec:eval-ft}). \sys{} makes three types of yes/no decisions with it (orange in Figure~\ref{fig:overview}), which the LLM would otherwise have to make:
\begin{itemize}
\item \emph{Relevance}: $D_{\mathrm{rel}}(q, p)$ decides whether a retrieved passage $p$ is relevant to a question $q$.
\item \emph{Direct answer}: $D_{\mathrm{direct}}(Q, R_w(Q, \mathcal{C}))$ decides whether the top $w$ passages retrieved for the original question, $R_w(Q, \mathcal{C})$, already contain all the evidence needed, so that $Q$ can be answered in one step.
\item \emph{Enough facts}: $D_{\mathrm{suff}}(Q, F_t)$ decides whether the QA pairs found so far are enough to answer $Q$, which ends the loop.
\end{itemize}
For each decision, the DM reads a short input $x$, such as a question and a passage, and returns the probability that the answer is yes. The decision is
\begin{equation}
  D(x) = \mathbb{1}\big[\,p_{\mathrm{DM}}(\text{yes} \mid x) \ge \tau\,\big],
  \label{eq:decision}
\end{equation}
with a fixed threshold $\tau$ for each type. Since the DM's probability of yes reflects how often yes is actually right (Appendix~\ref{app:training}), we use $\tau = 0.5$, ``more likely yes than no'', for relevance and enough facts, and a stricter $\tau = 0.7$ for direct answer, because a question that needs several steps but is wrongly answered directly is unlikely to get the right answer, while a question that one step could answer but is wrongly sent to the loop can still get the right answer there, only with more LLM calls.

\paragraph{Passage selection.} For relevance, the DM keeps, from the top $w$ passages retrieved for $q$, those relevant to $q$:
\begin{equation}
  \mathrm{Select}\big(q, R_w(q, \mathcal{C})\big) = \{\, p \in R_w(q, \mathcal{C}) : D_{\mathrm{rel}}(q, p) = 1 \,\}.
  \label{eq:select}
\end{equation}
Because only the selected passages reach the LLM, $w$ can be larger than the $k$ of the usual top-$k$ retrieval (Eq.~\eqref{eq:topk}) without burying the evidence among distracting passages (\S\ref{sec:bg-iter}). All settings are listed in Table~\ref{tab:settings}.

\paragraph{Fine-tuning.} We fine-tune the DM with LoRA~\cite{hu2022lora}, using labels derived only from the gold annotations of each benchmark's training split, such as which passages hold the evidence for a question and which intermediate answers it needs. No test question is used. Training details are given in Appendix~\ref{app:training}, and \S\ref{sec:eval-ftablation} shows that fine-tuning makes the decisions more accurate.

\subsection{Reusing Answers Across Questions}
\label{sec:design-reuse}
Because a self-contained sub-question names everything it asks about, it has the same meaning, and the same answer, whichever original question it was written for. For example, ``In which city is the headquarters of Sazerac Company located?'' has the answer ``New Orleans'' whether it arises from a question about Barton Premium Blend or about Blanton's, two whiskey brands made by that company. So a QA pair found while answering one question can be used again by any later question that needs it. \sys{} keeps the QA pairs of all questions in a QA cache $\mathcal{M}$ and checks it before retrieving passages for a new sub-question. It first looks for a stored sub-question with the same wording, ignoring letter case and punctuation. If none is found, it takes the few stored sub-questions most similar to the new one and uses the DM to decide whether any of them has the same meaning, with a strict $\tau = 0.8$, because a wrongly reused answer carries an error into another question. If one does, its answer is reused, without retrieval or an LLM call. Otherwise, the sub-question is answered as usual and the new QA pair is added to $\mathcal{M}$. In Figure~\ref{fig:overview}, the first sub-question was answered for an earlier question, so its answer comes from the QA cache (marked ``cached'').

\subsection{Query Answering Process}
\label{sec:design-process}
Algorithm~\ref{alg:deftrag} and Figure~\ref{fig:overview} show how the components work together to answer a question $Q$. \sys{} first retrieves $w$ passages for $Q$, and the DM decides whether they already contain all the evidence needed; if so, \lans{} answers $Q$ directly from the selected passages (lines 1--3). Otherwise, \sys{} starts with an empty QA table and proceeds one step at a time: \lsub{} writes the next self-contained sub-question, the DM selects the relevant passages retrieved for it, \lans{} answers it, and the pair is added to the QA table (lines 5--9). After each step, the DM checks whether the QA pairs are enough to answer $Q$; if so, \lfin{} answers $Q$ from the QA table (lines 10--11). If this does not happen within $T$ steps, \sys{} falls back to answering $Q$ directly from the relevant passages among the top $2w$ retrieved for it (lines 14--15).

\begin{algorithm}[t]
\caption{\sys{} query answering}
\label{alg:deftrag}
\begin{algorithmic}[1]
\REQUIRE question $Q$, corpus $\mathcal{C}$, number of retrieved passages $w$, step limit $T$
\STATE $P \gets \mathrm{Select}(Q, R_w(Q, \mathcal{C}))$
\IF{$D_{\mathrm{direct}}(Q, R_w(Q, \mathcal{C}))$}
  \RETURN $\lans(Q, P)$
\ENDIF
\STATE $F \gets \emptyset$
\FOR{$t = 1, \dots, T$}
  \STATE $q \gets \lsub(Q, F)$
  \STATE $a \gets \lans\big(q, \mathrm{Select}(q, R_w(q, \mathcal{C}))\big)$
  \STATE \algorithmicif\ $a \neq \varnothing$ \algorithmicthen\ $F \gets F \cup \{(q, a)\}$
  \IF{$D_{\mathrm{suff}}(Q, F)$}
    \RETURN $\lfin(Q, F)$
  \ENDIF
\ENDFOR
\STATE $P \gets \mathrm{Select}(Q, R_{2w}(Q, \mathcal{C}))$
\RETURN $\lans(Q, P)$
\end{algorithmic}
\end{algorithm}

The direct-answer decision lets questions that need only one retrieval be answered in one step, which saves LLM calls; for the other questions, the final answer is built from a short list of QA pairs rather than from a long reasoning trace.

\section{Evaluation}
\label{sec:eval}
\begin{table*}[t]
\centering
\small
\caption{Results on 1{,}000 questions per benchmark with two backbones: exact match (EM), token F1, and LLM-judged accuracy (Acc). gpt-4o-mini runs are judged by Qwen3.6 and Qwen3.6 runs by gpt-4o-mini, so no model grades its own answers. EM and F1 are computed from the answers as given and understate systems that answer in full sentences, such as LightRAG with gpt-4o-mini; Acc is our main metric. Avg: mean over the three benchmarks. Best in bold, second best underlined.}
\label{tab:main}
\begin{tabular*}{\textwidth}{@{\extracolsep{\fill}}l|ccc|ccc|ccc|ccc}
\toprule
\multirow{2}{*}{System} & \multicolumn{3}{c|}{MuSiQue} & \multicolumn{3}{c|}{2Wiki} & \multicolumn{3}{c|}{HotpotQA} & \multicolumn{3}{c}{Avg} \\
\cmidrule(lr){2-4}\cmidrule(lr){5-7}\cmidrule(lr){8-10}\cmidrule(lr){11-13}
 & EM & F1 & Acc & EM & F1 & Acc & EM & F1 & Acc & EM & F1 & Acc \\
\midrule
\multicolumn{13}{l}{\emph{gpt-4o-mini backbone}} \\
\midrule
No Retrieval & 0.122 & 0.226 & 0.239 & 0.273 & 0.351 & 0.410 & 0.268 & 0.397 & 0.527 & 0.221 & 0.325 & 0.392 \\
Vanilla RAG & 0.238 & 0.347 & 0.412 & 0.459 & 0.553 & 0.659 & 0.468 & 0.619 & 0.799 & 0.388 & 0.506 & 0.623 \\
\midrule
LightRAG & 0.000 & 0.087 & 0.305 & 0.000 & 0.150 & 0.505 & 0.000 & 0.190 & 0.696 & 0.000 & 0.142 & 0.502 \\
PathRAG & 0.123 & 0.233 & 0.276 & 0.278 & 0.379 & 0.451 & 0.397 & 0.549 & 0.693 & 0.266 & 0.387 & 0.473 \\
HippoRAG & 0.189 & 0.286 & 0.323 & 0.461 & 0.550 & 0.650 & 0.389 & 0.515 & 0.676 & 0.346 & 0.450 & 0.550 \\
HippoRAG 2 & 0.303 & 0.428 & 0.478 & 0.558 & 0.639 & 0.713 & 0.546 & 0.691 & 0.813 & 0.469 & 0.586 & 0.668 \\
MemGraphRAG & 0.275 & 0.398 & 0.453 & 0.524 & 0.576 & 0.635 & 0.514 & 0.665 & 0.795 & 0.438 & 0.546 & 0.628 \\
G3RAG & 0.327 & 0.450 & 0.503 & 0.542 & 0.611 & 0.682 & 0.563 & 0.711 & 0.830 & 0.477 & 0.590 & 0.672 \\
LinearRAG & 0.243 & 0.375 & 0.439 & 0.450 & 0.611 & 0.786 & 0.439 & 0.625 & 0.822 & 0.377 & 0.537 & 0.682 \\
\midrule
LogicRAG & 0.333 & 0.459 & 0.495 & 0.586 & 0.694 & 0.772 & 0.559 & 0.688 & 0.798 & 0.493 & 0.614 & 0.688 \\
IRCoT & 0.348 & 0.465 & 0.510 & 0.619 & 0.701 & 0.767 & 0.595 & 0.721 & 0.823 & 0.521 & 0.629 & 0.700 \\
Self-Ask & 0.339 & 0.455 & 0.500 & 0.589 & 0.695 & \underline{0.849} & 0.451 & 0.596 & 0.783 & 0.460 & 0.582 & 0.711 \\
Hi-Q & \underline{0.414} & \underline{0.549} & \underline{0.577} & \underline{0.631} & \underline{0.720} & 0.812 & \underline{0.600} & \underline{0.734} & \underline{0.840} & \underline{0.548} & \underline{0.667} & \underline{0.743} \\
\midrule
\textbf{\sys{} (ours)} & \textbf{0.453} & \textbf{0.587} & \textbf{0.629} & \textbf{0.678} & \textbf{0.767} & \textbf{0.852} & \textbf{0.627} & \textbf{0.775} & \textbf{0.874} & \textbf{0.586} & \textbf{0.709} & \textbf{0.785} \\
\midrule
\multicolumn{13}{l}{\emph{Qwen3.6-35B-A3B backbone}} \\
\midrule
No Retrieval & 0.128 & 0.216 & 0.198 & 0.366 & 0.417 & 0.428 & 0.293 & 0.392 & 0.427 & 0.262 & 0.341 & 0.351 \\
Vanilla RAG & 0.297 & 0.373 & 0.397 & 0.553 & 0.618 & 0.670 & 0.604 & 0.726 & 0.809 & 0.485 & 0.572 & 0.625 \\
\midrule
LightRAG & 0.270 & 0.385 & 0.459 & 0.367 & 0.493 & 0.688 & 0.400 & 0.598 & 0.816 & 0.346 & 0.492 & 0.654 \\
PathRAG & 0.333 & 0.439 & 0.444 & 0.524 & 0.607 & 0.667 & 0.599 & 0.727 & 0.813 & 0.485 & 0.591 & 0.641 \\
HippoRAG & 0.187 & 0.247 & 0.257 & 0.535 & 0.604 & 0.676 & 0.453 & 0.554 & 0.621 & 0.392 & 0.468 & 0.518 \\
HippoRAG 2 & 0.346 & 0.439 & 0.471 & 0.625 & 0.682 & 0.744 & 0.604 & 0.727 & 0.818 & 0.525 & 0.616 & 0.678 \\
MemGraphRAG & 0.265 & 0.340 & 0.388 & 0.537 & 0.576 & 0.621 & 0.542 & 0.654 & 0.740 & 0.448 & 0.524 & 0.583 \\
G3RAG & 0.411 & 0.501 & 0.534 & 0.622 & 0.679 & 0.725 & 0.633 & 0.752 & 0.839 & 0.555 & 0.644 & 0.699 \\
LinearRAG & 0.353 & 0.437 & 0.475 & 0.671 & 0.758 & 0.830 & 0.606 & 0.746 & 0.857 & 0.543 & 0.647 & 0.721 \\
\midrule
LogicRAG & 0.391 & 0.503 & 0.529 & \underline{0.696} & \underline{0.779} & \underline{0.838} & 0.620 & 0.745 & 0.830 & 0.569 & 0.675 & 0.732 \\
IRCoT & 0.383 & 0.482 & 0.486 & 0.678 & 0.747 & 0.799 & 0.629 & 0.761 & 0.843 & 0.563 & 0.663 & 0.709 \\
Self-Ask & 0.423 & 0.526 & 0.546 & 0.677 & 0.762 & 0.825 & 0.581 & 0.708 & 0.795 & 0.560 & 0.665 & 0.722 \\
Hi-Q & \underline{0.466} & \underline{0.574} & \underline{0.583} & 0.600 & 0.681 & 0.783 & \underline{0.652} & \underline{0.776} & \underline{0.864} & \underline{0.573} & \underline{0.677} & \underline{0.743} \\
\midrule
\textbf{\sys{} (ours)} & \textbf{0.494} & \textbf{0.615} & \textbf{0.636} & \textbf{0.716} & \textbf{0.801} & \textbf{0.867} & \textbf{0.658} & \textbf{0.799} & \textbf{0.871} & \textbf{0.623} & \textbf{0.738} & \textbf{0.791} \\
\bottomrule
\end{tabular*}
\end{table*}

\subsection{Setup}
\label{sec:eval-setup}
We evaluate on three multi-hop QA benchmarks, MuSiQue~\cite{trivedi2022musique}, 2WikiMultiHopQA (2Wiki)~\cite{ho2020constructing} and HotpotQA~\cite{yang2018hotpotqa}, using the 1{,}000-question test set of each and its passage corpus (11{,}656, 6{,}119 and 9{,}811 passages), as released by the authors of LogicRAG~\cite{chen2026logicrag}. We compare against 13 baselines: answering without retrieval (No Retrieval) and vanilla RAG; seven graph-based methods, LightRAG~\cite{guo2024lightrag}, PathRAG~\cite{chen2025pathrag}, HippoRAG~\cite{gutierrez2024hipporag}, HippoRAG~2~\cite{gutierrez2025hipporag2}, MemGraphRAG~\cite{wu2026memgraphrag}, G3RAG~\cite{li2026g3rag} and LinearRAG~\cite{zhuang2025linearrag}; and four iterative methods, IRCoT~\cite{trivedi2023ircot}, Self-Ask~\cite{press2023selfask}, LogicRAG~\cite{chen2026logicrag} and Hi-Q~\cite{kim2026hiq}. We do not compare with REAP~\cite{zhu2026reap}, whose planner and fact extractor are LLMs fine-tuned on trajectories from HotpotQA and 2Wiki, which does not fit our setting of one shared backbone for all systems. All systems share the same backbone LLM, the same embedding model (Qwen3-Embedding-4B), temperature~0 and the same output limit; each baseline otherwise keeps its own published settings, such as chunk size.

We use two backbones: gpt-4o-mini, through its API, and Qwen3.6-35B-A3B, run locally. All baselines retrieve the top $k = 5$ passages per retrieval. \sys{} retrieves the top $w = 2k = 10$ passages, from which the DM keeps the relevant ones. For a fair comparison, it keeps at most five relevant passages at each step (2.1 to 2.4 on average), the same number as the baselines' top-5. \sys{} takes at most $T = 6$ steps, and all main results are obtained without reuse (\S\ref{sec:design-reuse}). We report exact match (EM), token F1 and LLM-judged accuracy (LLM-Acc), our main metric, graded with the SimpleQA prompt~\cite{wei2024simpleqa}. To avoid a model grading its own answers, gpt-4o-mini runs are judged by Qwen3.6 and Qwen3.6 runs by gpt-4o-mini. Differences are tested with the exact McNemar test on paired answers. Appendix~\ref{app:overhead} reports the overhead of every method: index building, answering latency, LLM calls and tokens.

\subsection{Fine-Tuning the Decision Model}
\label{sec:eval-ft}
The DM starts from the StartLux-Decision-4B weights~\cite{startlux2026} (Qwen3.5 architecture, 4B parameters), which were trained to answer yes/no and multiple-choice questions with probabilities that reflect how often the answer is right. We fine-tune it on data built only from the \emph{training} splits of the three benchmarks. To make the training inputs look like the test inputs, we split the training questions into groups of 1{,}000, and each group retrieves from a corpus made of its own passages, just as each test set retrieves from its own corpus. The labels come from the benchmarks' gold annotations: a passage is relevant if it is a gold supporting passage; a question can be answered directly if all its supporting passages are retrieved; and the QA pairs found so far are enough once they include the answer to every step of the question's gold decomposition (MuSiQue) or gold evidence (2Wiki). No test question or test-corpus passage appears in the training data, which has 48{,}036 training and 7{,}806 development examples (the latter from 15\% of the training questions, held out from training). We fine-tune with LoRA~\cite{hu2022lora} (rank 16, alpha 32, dropout 0.05, on all linear layers; 32.5M of 4.24B parameters trained), at a learning rate of $10^{-4}$ with 5\% warm-up and cosine decay, for one epoch (2{,}262 steps, about 6.4 hours on two NVIDIA RTX A5000 GPUs). We then adjust the output probabilities with a single temperature parameter fitted on the development examples, never on test questions. Appendix~\ref{app:training} gives the full data recipe.

\subsection{Main Results}
\label{sec:eval-main}

Table~\ref{tab:main} reports EM, F1 and accuracy on all three benchmarks with both backbones. \sys{} achieves the highest score on all three metrics in all six settings; we discuss accuracy, our main metric, below. Its gains over the second-best system are largest on MuSiQue, which has the longest evidence chains: $+5.2$ points over Hi-Q with gpt-4o-mini (0.629 vs.\ 0.577) and $+5.3$ with Qwen3.6 (0.636 vs.\ 0.583), both significant. On 2Wiki, \sys{} is 2.9 points ahead of LogicRAG with Qwen3.6 (significant) and on par with Self-Ask with gpt-4o-mini (0.852 vs.\ 0.849). On HotpotQA, it is 3.4 points ahead of Hi-Q with gpt-4o-mini (significant) and on par with Hi-Q and LinearRAG with Qwen3.6 (0.871 vs.\ 0.864 and 0.857). Against every other baseline, the difference is significant in every setting.

Graph-based methods vary widely: with gpt-4o-mini, LightRAG, PathRAG and HippoRAG fall below vanilla RAG, and even the strongest of them trail \sys{} by at least 10 points on MuSiQue with either backbone, which matches their low full-chain coverage (\S\ref{sec:bg-graph}). The strongest baselines are iterative methods, but the best of them changes with the benchmark and the backbone: Hi-Q on MuSiQue and HotpotQA, and Self-Ask or LogicRAG on 2Wiki. \sys{} is the only system that is first or on par with the first everywhere.

\subsection{Why \sys{} Gives Better Answers}
\label{sec:eval-why}
\begin{table}[t]
\centering
\small
\caption{Evidence coverage on MuSiQue (gpt-4o-mini), for questions that need 2, 3 or 4 hops: the share of questions for which the LLM received all passages needed to answer them, as annotated in the dataset. Words: the average number of words of passages the LLM received per question.}
\label{tab:coverage-full}
\begin{tabular}{lcccr}
\toprule
Method & 2-hop & 3-hop & 4-hop & Words \\
\midrule
Vanilla RAG & 46\% & 7\% & 1\% & 870 \\
\midrule
\multicolumn{5}{l}{\emph{Graph-based}} \\
PathRAG & 42\% & 10\% & 8\% & 12,390 \\
HippoRAG & 33\% & 7\% & 0\% & 520 \\
HippoRAG 2 & 53\% & 8\% & 0\% & 500 \\
MemGraphRAG & 41\% & 7\% & 1\% & 500 \\
G3RAG & 48\% & 6\% & 0\% & 470 \\
LinearRAG & 61\% & 28\% & 19\% & 4,110 \\
\midrule
\multicolumn{5}{l}{\emph{Iterative}} \\
IRCoT & 76\% & 37\% & 7\% & 780 \\
Self-Ask & 83\% & 41\% & 14\% & 2,590 \\
LogicRAG & 59\% & 19\% & 2\% & 750 \\
Hi-Q & 72\% & 28\% & 4\% & 920 \\
\midrule
\textbf{\sys{}} & 76\% & 46\% & 18\% & 390 \\
\bottomrule
\end{tabular}
\end{table}

\paragraph{More of the evidence chain, in fewer words.} Table~\ref{tab:coverage-full} measures, for each MuSiQue question, whether the LLM received all passages needed to answer it, as annotated in the dataset, and how many words of passages it received. \sys{} gives the LLM the full chain for 46\% of 3-hop and 18\% of 4-hop questions, the highest 3-hop coverage of all systems, with 390 words per question, the fewest of any system and less than half of vanilla RAG's 870. The systems that cover more chains on some hop count use far more words: Self-Ask covers more 2-hop chains (83\%) with 2{,}590 words, and LinearRAG covers 19\% of 4-hop chains with 4{,}110 words, about ten times as many as \sys{}. These extra words bury the needed passages among many that are not needed, which lowers accuracy (\S\ref{sec:bg-iter}, Table~\ref{tab:window}), and they add to the LLM's input cost. Asking one self-contained sub-question at a time and keeping only the passages the DM judges relevant gives the LLM a short context that still contains the needed evidence.

\begin{table}[t]
\centering
\small
\setlength{\tabcolsep}{4.5pt}
\caption{Effect of selecting relevant passages: \sys{} versus the same process without selection (the top-5 retrieved passages, all kept). gpt-4o-mini backbone, 300 test questions per benchmark (a subset of those in Table~\ref{tab:main}); Avg is the mean over the three benchmarks.}
\label{tab:selection}
\begin{tabular}{lcccc}
\toprule
& MuSiQue & 2Wiki & HotpotQA & Avg \\
\midrule
\sys{} (top-10, selected) & 0.633 & 0.860 & 0.880 & 0.791 \\
No selection (top-5) & 0.610 & 0.823 & 0.827 & 0.753 \\
\midrule
Difference & +2.3 & +3.7 & +5.3 & +3.8 \\
\bottomrule
\end{tabular}
\end{table}

\paragraph{Effect of passage selection.} To isolate the effect of selection, we turn it off and give the LLM the top-5 retrieved passages, as the baselines do, keeping everything else the same (Table~\ref{tab:selection}). Without selection, accuracy drops on all three benchmarks, by 2.3, 3.7 and 5.3 points on MuSiQue, 2Wiki and HotpotQA; the drop on HotpotQA is significant ($p = 0.002$). On 2Wiki and HotpotQA, most of the loss comes from questions answered directly, where the LLM must find the evidence for the whole question in one list of passages. This agrees with Table~\ref{tab:window}: on questions that can be answered in one retrieval step, a longer list of passages alone barely helps, but keeping only the needed passages from it raises accuracy substantially.

\subsection{Effect of Fine-Tuning the Decision Model}
\label{sec:eval-ftablation}
\begin{table}[t]
\centering
\small
\caption{Effect of fine-tuning on decision quality, base DM / fine-tuned DM, on 300 test questions per benchmark. AUC measures how well the DM's yes probabilities separate yes cases from no cases (1.0 is perfect). HotpotQA has no answer for each step, so its enough-facts label is yes when the passages used include all gold supporting passages (167 cases).}
\label{tab:finetune}
\begin{tabular}{lccc}
\toprule
Decision & MuSiQue & 2Wiki & HotpotQA \\
\midrule
Relevance & 0.931 / \textbf{0.979} & 0.996 / \textbf{1.000} & 0.990 / \textbf{0.997} \\
Direct answer & 0.850 / \textbf{0.933} & 0.938 / \textbf{0.997} & 0.915 / \textbf{0.995} \\
Enough facts & 0.903 / \textbf{0.904} & 0.943 / \textbf{0.959} & 0.816 / \textbf{0.845} \\
\bottomrule
\end{tabular}
\end{table}

Table~\ref{tab:finetune} compares the decisions of the base DM (StartLux-Decision-4B before fine-tuning) and the fine-tuned DM against gold labels. Fine-tuning improves the direct-answer decision the most, for example from 0.850 to 0.933 on MuSiQue, and also improves relevance, most clearly on MuSiQue (0.931 to 0.979), while the enough-facts decision changes little. End to end, on the first 300 MuSiQue questions with gpt-4o-mini, the fine-tuned DM raises accuracy from 0.587 to 0.623.

\subsection{Reusing Answers}
\label{sec:eval-reuse}
We run \sys{} with reuse (\S\ref{sec:design-reuse}) on all 1{,}000 MuSiQue questions with gpt-4o-mini, answering them in order with one QA cache. Of all sub-questions generated over the 1{,}000 questions, 23.1\% are answered from the cache, which reduces the number of LLM calls by 12.9\%. Accuracy does not drop (0.642 with reuse vs.\ 0.629 without, $p = 0.18$), so answers found for earlier questions can be safely reused for later ones.

\section{Related Work}
\label{sec:related}

\paragraph{Graph-based RAG\@.} Graph-based methods build a graph over the corpus offline and retrieve along it at answering time. GraphRAG~\cite{edge2024graphrag} and LightRAG~\cite{guo2024lightrag} extract entities and relations with an LLM and retrieve community summaries or entities, PathRAG~\cite{chen2025pathrag} retrieves relational paths, HippoRAG~\cite{gutierrez2024hipporag} and HippoRAG~2~\cite{gutierrez2025hipporag2} rank passages with Personalized PageRank from the question's entities, MemGraphRAG~\cite{wu2026memgraphrag} builds a layered graph of schemas, facts and passages, and LinearRAG~\cite{zhuang2025linearrag} and G3RAG~\cite{li2026g3rag} build their graphs without LLM extraction. All of these methods retrieve once per question, starting from what the question names, which rarely recovers the full evidence chain beyond two hops (\S\ref{sec:bg-graph}). \sys{} instead finds the bridge entities at answering time, one self-contained sub-question at a time, and retrieves the full chain for more questions with three or more hops than any of these methods (Table~\ref{tab:coverage-full}).

\paragraph{Iterative RAG\@.} IRCoT~\cite{trivedi2023ircot} uses each new chain-of-thought sentence as the next query, and Self-Ask~\cite{press2023selfask} poses follow-up questions within a single growing prompt. LogicRAG~\cite{chen2026logicrag} and REAP~\cite{zhu2026reap} plan the sub-questions with their dependencies in advance, and Hi-Q~\cite{kim2026hiq} and REAP rewrite dependent sub-questions as answers arrive. In these methods, the LLM itself takes on several tasks, often within the same call: planning or writing the sub-questions, deciding their dependencies, rewriting them, keeping track of what is known, and deciding when to stop. \sys{} instead lets the LLM focus on a single fine-grained task at a time: generating one self-contained sub-question for the next missing piece of information, which removes planning, dependencies and rewriting, and avoids their associated errors; tracking is handled by the QA table, and the yes/no decisions by the fine-tuned DM. Its LLM input stays short, 1{,}569 tokens per question against 6{,}140 for Self-Ask (Appendix~\ref{app:overhead}), and it achieves the highest accuracy of these methods in every setting (Table~\ref{tab:main}).

\paragraph{Models for retrieval decisions.} Rerankers~\cite{nogueira2020monot5,sun2023rankgpt} re-score retrieved passages but still keep a fixed number, and Adaptive-RAG~\cite{jeong2024adaptiverag} uses a trained classifier to choose between no, single and iterative retrieval; Self-RAG~\cite{asai2024selfrag}, Search-R1~\cite{jin2025searchr1}, R1-Searcher~\cite{song2025r1searcher} and Search-o1~\cite{li2025searcho1} instead train the backbone LLM, or use a reasoning model, to make such decisions. \sys{} uses one fine-tuned DM for the yes/no decisions at every step while the backbone stays unchanged, so any LLM can serve as the backbone, and fine-tuning the DM is much cheaper than training the backbone LLM.

\section{Conclusion}
\label{sec:conclusion}
We presented \sys{}, a RAG framework for multi-hop question answering built on three components: incremental decomposition into self-contained sub-questions, which avoids planning, dependency and rewriting errors; fine-grained tasks handled outside the LLM, namely rule-based updates to a QA table and yes/no decisions by a fine-tuned lightweight DM; and a QA cache that reuses answers across questions. Across three benchmarks and two backbones, \sys{} achieves the best accuracy among thirteen baselines, with its largest gains on MuSiQue, while giving the LLM fewer input tokens than any other iterative method; reusing answers reduces LLM calls without lowering accuracy.

\bibliographystyle{ACM-Reference-Format}
\bibliography{refs}

\appendix
\section{Implementation Details}
\label{app:algorithm}
Table~\ref{tab:settings} lists the settings used for all results. Beyond Algorithm~\ref{alg:deftrag}, the implementation retries once on an empty or repeated reply, with a changed input, and uses the following rules.

\paragraph{Additional rules.} A question is answered directly only when the DM judges its passages sufficient ($\tau = 0.7$) and it does not ask for a composite family relation such as a grandparent or an in-law, which is detected by keywords. All results use the same rules on every benchmark and backbone.

\paragraph{How the thresholds were chosen.} The thresholds of 0.5 are the natural cut-off for yes/no probabilities that reflect how often yes is right, and were not tuned. The direct-answer threshold of 0.7 was chosen during development on stored answers of part of the test questions (2Wiki and the first 300 MuSiQue questions), and then kept fixed for all benchmarks and both backbones. The result depends little on this choice: moving the threshold to 0.6 or 0.8 changes the path of at most 13 of the 1{,}000 questions per benchmark and accuracy by at most 0.1 points on every benchmark (gpt-4o-mini). The reuse threshold of 0.8 was set strict by hand, because a wrongly reused answer carries an error into another question.

\begin{table}[h]
\centering
\small
\caption{Settings of \sys{} used for all results.}
\label{tab:settings}
\begin{tabular}{lr}
\toprule
Setting & Value \\
\midrule
Retrieved passages $w$ (retry and fallback) & 10 (20) \\
Step limit $T$ & 6 \\
Relevance: keep if $p(\text{yes}) \ge$ & 0.5 \\
Passages kept when only one / no passage is judged relevant & 2 / 5 \\
Passages kept at most & 5 \\
Direct answer: sufficient $\ge$ & 0.7 \\
Enough facts: $p(\text{yes}) \ge$ & 0.5 \\
Reuse: stored sub-questions compared & 5 \\
Reuse: same meaning $p(\text{yes}) \ge$ & 0.8 \\
Passage length cap (characters) & 3{,}000 \\
\bottomrule
\end{tabular}
\end{table}

\section{Prompts}
\label{app:prompts}
The exact prompts of the three LLM tasks and the yes/no questions asked to the DM. Text in braces is filled in at run time. The prompts call the QA pairs ``facts''.

\paragraph{Writing a sub-question, no QA pairs yet ($\lsub$).} {\scriptsize\ttfamily\raggedright A question may refer to something by description instead of by name, such as \textquotedbl{}the author of Y\textquotedbl{}. Some descriptions are built on others: \textquotedbl{}the birthplace of the author of Y\textquotedbl{} needs \textquotedbl{}the author of Y\textquotedbl{} first.\par\smallskip Write ONE short question that finds a piece of information needed to answer the question below. Ask about a description that is not built on another one, so your question can be searched on its own. If the question has no such description and needs no calculation, ask the question itself.\par\smallskip Example\\
Question: What language is spoken in the birthplace of the author of Y?\\
Next question: Who is the author of Y?\par\smallskip Once the people or things the question describes are known by name, ask for a fact a document states directly, of the kind the question asks about: to find out who is older, ask for each person\textquotesingle{}s date of birth, not how old they are; to find out which country someone is from, ask for their nationality, not their place of birth.\par\smallskip Question: \{question\}\par\smallskip Reply with only the next question.\par}

\paragraph{Writing a sub-question, with QA pairs ($\lsub$).} {\scriptsize\ttfamily\raggedright Below is a question, and the facts found so far while working on it.\par\smallskip A question may refer to something by description instead of by name, such as \textquotedbl{}the author of Y\textquotedbl{}. Some descriptions are built on others: \textquotedbl{}the birthplace of the author of Y\textquotedbl{} needs \textquotedbl{}the author of Y\textquotedbl{} first.\par\smallskip Write ONE short question that finds the next missing piece of information needed to answer the main question. Use the facts found so far: write names in place of descriptions the facts already resolve. Ask about a description that is not built on another unresolved one, so your question can be searched on its own. Do not ask for anything already in the facts.\par\smallskip Example\\
Question: What language is spoken in the birthplace of the author of Y?\\
Facts found so far:\\
1. Who is the author of Y? Answer: Jane Doe\\
Next question: Where was Jane Doe born?\par\smallskip Once the people or things the question describes are known by name, ask for a fact a document states directly, of the kind the question asks about: to find out who is older, ask for each person\textquotesingle{}s date of birth, not how old they are; to find out which country someone is from, ask for their nationality, not their place of birth.\par\smallskip Question: \{question\}\\
Facts found so far:\\
\{facts\}\par\smallskip
Reply with only the next question.\par}

\paragraph{Added after a sub-question got no answer.} {\scriptsize\ttfamily\raggedright These questions could not be answered from the documents:\\
\{failed\}\\
Do not ask any of these again. Ask for the needed information in different words, or work on another part of the main question.\par}

\paragraph{Added after a repeated sub-question.} {\scriptsize\ttfamily\raggedright The question \textquotedbl{}\{sq\}\textquotedbl{} was already asked. Write a different question.\par}

\paragraph{Answering from passages ($\lans$).} {\scriptsize\ttfamily\raggedright Answer the question using ONLY the passages below. Give a short, direct answer -{}- a few words, not a sentence.\par\smallskip Passages:\\
\{passages\}\par\smallskip
Question: \{question\}\\
Answer:\par}

\paragraph{Answering the original question from the QA table ($\lfin$).} {\scriptsize\ttfamily\raggedright Answer the question using ONLY the facts below. Give a short, direct answer -{}- a few words, not a sentence.\par\smallskip When the question compares dates, compare the years first, then the months, then the days. When it compares ages, the person born earlier is older.\par\smallskip Facts:\\
\{facts\}\par\smallskip
Question: \{question\}\\
Answer:\par}

\paragraph{Decision: relevance of passage $i$.} {\scriptsize\ttfamily\raggedright Is \textasciigrave{}p\_i\textasciigrave{} needed to answer \textasciigrave{}question\textasciigrave{}: does it state part of the answer, or a linking fact the reasoning must pass through (e.g., identifying an entity the question refers to only by description)?\\
Yes: \textasciigrave{}p\_i\textasciigrave{} states part of the answer, or a linking fact needed to reach it.\\
No: \textasciigrave{}p\_i\textasciigrave{} is not needed to answer \textasciigrave{}question\textasciigrave{}: it is unrelated or only loosely related.\par}

\paragraph{Decision: direct answer.} {\scriptsize\ttfamily\raggedright Do the passages \textasciigrave{}p1\textasciigrave{} through \textasciigrave{}p10\textasciigrave{} together contain all the facts needed to answer \textasciigrave{}question\textasciigrave{} -{}- whether stated outright or derivable by combining facts across them?\\
Yes: Every fact needed is present across the passages; the answer is stated or derivable by connecting them.\\
No: At least one fact needed to answer is missing and would have to be found elsewhere.\par}

\paragraph{Decision: enough facts.} {\scriptsize\ttfamily\raggedright Has \textasciigrave{}question\textasciigrave{} already been answered by the facts in \textasciigrave{}facts\textasciigrave{}: does every entity or value the question asks for, or depends on, appear among the facts\textquotesingle{} answers?\\
Yes: The facts already give the answer: each step the question depends on is answered in \textasciigrave{}facts\textasciigrave{}.\\
No: Some step the question depends on has not been answered yet, so \textasciigrave{}question\textasciigrave{} cannot be answered from \textasciigrave{}facts\textasciigrave{} alone.\par}

\paragraph{Decision: same meaning (reuse).} {\scriptsize\ttfamily\raggedright Are \textasciigrave{}query\textasciigrave{} and \textasciigrave{}c\_i\textasciigrave{} the same question: asking for the same piece of information about the same explicitly named entity, so that one correct answer is correct for both?\\
Yes: Both name the same entity explicitly and ask for the same information; only the wording differs.\\
No: They differ in entity, property, time, place or scope -{}- OR either one refers to something without naming it (such as \textquotedbl{}the tournament\textquotedbl{} or \textquotedbl{}the author\textquotedbl{}), so it is unclear whether they mean the same thing.\par}

\section{Decision Model Fine-Tuning Details}
\label{app:training}
\paragraph{Training questions and corpora.} For each benchmark, we sample 3{,}000 questions from its training split, following the mix of hop counts or question types of the test set, and split them into groups of 1{,}000. Each group's corpus consists of the passages of its own questions (MuSiQue about 13{,}000, 2Wiki about 5{,}200, HotpotQA about 9{,}600 passages per group, close to the test corpora of 11{,}656, 6{,}119 and 9{,}811). Every training request retrieves its top 10 passages from its group's corpus with the embedder and query instruction of \sys{}.

\paragraph{Labels.} All labels come from the training splits' gold annotations. \emph{Relevance}: a passage is relevant if it is a gold supporting passage; each request labels all gold passages and three randomly chosen other passages, while all ten passages stay in the request. \emph{Direct answer}: yes if all gold supporting passages are among the top 10. \emph{Enough facts}: built from the gold decompositions (MuSiQue) and gold evidence (2Wiki): the answers to only the first steps give ``no'', the answers to all steps give ``yes'', a wrong answer to a step gives ``no'', and the answers to all steps plus an unrelated QA pair give ``yes''. We add enough-facts items taken from runs of \sys{} on training questions, labelled by the same rule, and items whose QA pairs use self-contained sub-questions, as \sys{} writes them. The same-meaning decision is not fine-tuned.

\paragraph{Leakage checks.} All 1{,}000 test questions of each benchmark come from its official validation split; none of them is in the training data, and no training question text is identical to a test question. Training questions whose gold passage appears in any test corpus are skipped, so that no test-corpus passage is ever shown in training.

\paragraph{Training and probability adjustment.} The final data has 48{,}036 training and 7{,}806 development examples. LoRA settings are given in \S\ref{sec:eval-ft}; training used micro-batches of 8{,}192 tokens, gradient accumulation to an effective batch equal to the base model's single-GPU default, gradient checkpointing and gradient clipping at 1.0. A single temperature for yes/no probabilities is fitted on the development examples (1.08).

\section{Experimental Details for Baselines}
\label{app:baselines}
All systems use the same backbone, the same embedding model (Qwen3-Embedding-4B, last-token pooling, with the query instruction applied to queries only), temperature~0, $\text{top-}p = 1$ and the same output limit of 3{,}072 tokens; the reasoning mode of Qwen3.6 is turned off for every system. Each baseline keeps the settings of its released code or paper that define the method, such as chunk size and graph construction, and all baselines retrieve the top $k = 5$ passages per retrieval (\S\ref{sec:eval-setup}). Systems that build the same kind of index share it: HippoRAG and MemGraphRAG reuse the index built by HippoRAG~2 (for MemGraphRAG, only the passage embeddings, after we checked that the passage sets are identical), and PathRAG reuses LightRAG's graph. Every system's answers are graded by the same script with the same judge prompt, and missing answers count as wrong.

\section{Overhead Analysis}
\label{app:overhead}
Table~\ref{tab:overhead-all} reports, for every method with gpt-4o-mini, the cost of building the index and the cost of answering each question. Its last row replaces the local DM of \sys{} with Jev, a closed decision model accessed through an API.

\begin{table*}[t]
\centering
\small
\caption{Overhead of every method with gpt-4o-mini, mean over the three benchmarks (1{,}000 questions each). Index: LLM calls, LLM tokens (input + output) and build time, per corpus. Answering: per question, from question to answer (grading excluded). Methods with an index of passage embeddings only use no LLM to build it; their build time is that of embedding the corpus with Qwen3-Embedding-4B. $^a$Uses LightRAG's graph. $^b$Uses HippoRAG~2's graph. $^c$Includes the extraction it shares with HippoRAG~2. $^d$Build time not recorded; no LLM is used. $^e$Mean latency (only totals are recorded). Latency is the median over questions, because a few API stalls of several minutes dominate the mean. $^f$\sys{} with the closed decision model Jev, accessed through an API, in place of the local DM; its time per call is set to its typical 0.16~s.}
\label{tab:overhead-all}
\begin{tabular*}{\textwidth}{@{\extracolsep{\fill}}l|rrr|rrrr}
\toprule
 & \multicolumn{3}{c|}{Index (per corpus)} & \multicolumn{4}{c}{Answering (per question)} \\
Method & LLM calls & Tokens (M) & Time (h) & Latency (s) & LLM calls & Input tokens & Output tokens \\
\midrule
No Retrieval & 0 & 0 & 0.00 & 2.4 & 1.0 & 65 & 153 \\
Vanilla RAG & 0 & 0 & 0.09 & 2.4 & 1.0 & 1,279 & 142 \\
\midrule
LightRAG & 2,107 & 10.9 & 1.8 & 5.3 & 2.0 & 34,324 & 85 \\
PathRAG$^a$ & 2,107 & 10.9 & 1.8 & 3.8 & 2.0 & 17,291 & 47 \\
HippoRAG$^b$ & 18,391 & 10.7 & 0.8 & 3.4 & 2.0 & 956 & 150 \\
HippoRAG 2 & 18,391 & 10.7 & 0.8 & 3.9 & 2.0 & 4,250 & 147 \\
MemGraphRAG$^{c,e}$ & 146,555 & 114.3 & 17.9 & 1.7 & 1.0 & 1,389 & 90 \\
G3RAG & 0 & 0 & 0.11 & 2.9 & 2.0 & 3,692 & 124 \\
LinearRAG$^d$ & 0 & 0 & -- & 2.5 & 1.0 & 5,534 & 133 \\
\midrule
LogicRAG & 0 & 0 & 0.09 & 5.3 & 4.1 & 1,926 & 304 \\
IRCoT & 0 & 0 & 0.09 & 4.6 & 3.0 & 9,775 & 240 \\
Self-Ask & 0 & 0 & 0.09 & 6.8 & 6.3 & 6,140 & 308 \\
Hi-Q & 0 & 0 & 0.09 & 6.4 & 6.6 & 13,928 & 635 \\
\midrule
\textbf{\sys{}} & 0 & 0 & 0.09 & 9.9 & 4.8 & 1,569 & 33 \\
\sys{} + Jev$^f$ & 0 & 0 & 0.09 & 5.0 & 4.8 & 1,699 & 32 \\
\bottomrule
\end{tabular*}
\end{table*}

\paragraph{Index.} \sys{}, vanilla RAG and the iterative methods only embed the passages, which takes about five minutes per corpus and no LLM calls. The graph-based methods that extract a graph with the LLM need from about 2{,}100 (LightRAG) to 147{,}000 (MemGraphRAG) LLM calls and 11 to 114 million tokens per corpus, and up to 18 hours of build time; this cost is paid again when the corpus changes.

\paragraph{Answering.} \sys{} gives the LLM fewer input tokens per question than every other iterative method (1{,}569, against 1{,}926 to 13{,}928), and it produces the fewest output tokens of all methods (33), because each task asks for a short output; it also makes fewer LLM calls than Self-Ask and Hi-Q (4.8, against 6.3 and 6.6). Its latency is the highest (median 9.9~s).

\paragraph{Where the latency comes from.} Most of the answering time of \sys{} goes to its frequent calls to the local DM, which selects the relevant passages at every step and makes the other decisions. Per question, \sys{} makes 7.0, 6.3 and 3.2 DM calls on MuSiQue, 2Wiki and HotpotQA, at 1.6 to 1.8~s each, as the DM runs on one GPU, one request at a time, over inputs of up to ten passages. These calls take 12.5 of the 20.4~s mean answering time on MuSiQue, 10.3 of 16.7~s on 2Wiki and 5.2 of 7.8~s on HotpotQA, about 60 to 70\%. With Jev in place of the DM, at about 0.16~s per call, the median latency falls from 9.9 to 5.0~s, and the mean answering time to 8.2, 7.7 and 3.6~s on the three benchmarks, while accuracy stays within 1.2 points of the DM's (0.618, 0.857 and 0.867, against 0.629, 0.852 and 0.874). Most of the added latency thus comes from how the DM is served, not from the method itself.

\section{Case Studies}
\label{app:cases}
All cases are taken from the runs in Table~\ref{tab:main} (gpt-4o-mini backbone); each line shows a sub-question and the answer \sys{} found for it.

\paragraph{Where it works: a three-hop chain.} \emph{``When did the city where the headquarters of Barton Premium Blend's manufacturer is located elect its first black mayor?''} (gold: 1970s). \sys{} asks one self-contained sub-question at a time:
\begin{quote}\small
What is the name of the manufacturer of Barton Premium Blend? $\rightarrow$ Sazerac Company\\
In which city is the headquarters of Sazerac Company located? $\rightarrow$ New Orleans\\
When did New Orleans elect its first black mayor? $\rightarrow$ 1970s
\end{quote}
After the third QA pair, the DM judges the QA pairs enough and \lfin{} answers ``1970s''. Self-Ask's first follow-up skips the manufacturer (``What city is the headquarters of Barton Premium Blend's manufacturer located in?''), takes Bardstown, Kentucky as the answer, and finally answers ``Information not available''; Hi-Q also gives no answer, and LogicRAG answers ``No''.

\paragraph{Where it works, but not in order.} For ``Who sent naval ships to the body of water that joins the Atlantic and the sea where the Rhine ends?'' (gold: Caesar), the question behind the example in \S\ref{sec:design-units}, the DM first judges the retrieved passages insufficient (0.15), and \sys{} asks:
\begin{quote}\small
Who sent naval ships to the body of water that joins the Atlantic? $\rightarrow$ Caesar\\
What is the name of the sea where the Rhine ends? $\rightarrow$ North Sea\\
What is the name of the body of water that joins the Atlantic and the North Sea? $\rightarrow$ English Channel
\end{quote}
The first sub-question skips ahead to the last hop with a vague description and happens to retrieve the right answer. The DM does not stop there (enough facts: 0.44), but only after the chain through the North Sea and the English Channel is complete (0.70); the final answer is ``Caesar''.

\paragraph{Where it fails: nested sub-questions.} Bridge-comparison questions in 2Wiki compare two films through their directors. \emph{``Which film has the director who is older than the other, Airheads or Return To Cabin By The Lake?''} (gold: Return To Cabin By The Lake):
\begin{quote}\small
Who is the director of Airheads? $\rightarrow$ Michael Lehmann\\
What is the date of birth of Michael Lehmann? $\rightarrow$ March 30, 1957\\
What is the date of birth of the director of Return To Cabin By The Lake? $\rightarrow$ (no answer)\\
What is the name of the director of Return To Cabin By The Lake? $\rightarrow$ Po-Chih Leong\\
What is the date of birth of Po-Chih Leong? $\rightarrow$ 31 December 1939\\
What film did Po-Chih Leong direct? $\rightarrow$ (no answer)
\end{quote}
The third sub-question is not self-contained: it asks for the birth date of a director who has not been named yet, and retrieval finds no answer. \sys{} recovers in the next two steps, but the DM does not judge the four QA pairs enough, the loop reaches its step limit, and the fallback answers ``Airheads''. On 2Wiki bridge-comparison questions, 38\% of such nested sub-questions get no answer, against 4\% of sub-questions that ask for a single relation, and 75 of 235 of these questions reach the step limit or repeat themselves; this is the main reason \sys{} only matches Self-Ask on 2Wiki with gpt-4o-mini (Table~\ref{tab:main}).

\paragraph{Where it fails: the right entity, the wrong fact.} \emph{``When was the person who Messi's goals in Copa del Rey compared to get signed by Barcelona?''} (gold: June 1982). \sys{} first finds the right person (``Diego Maradona''), so it does not shift back to Messi; but it then asks for Maradona's date of birth instead of the date he was signed, the DM accepts these QA pairs as enough, and the final answer (``30 October 1984'') is wrong. The rule that tells \lsub{} to ask for dates of birth when comparing ages (Appendix~\ref{app:prompts}) may have pulled the sub-question towards a birth date.

\paragraph{Remaining errors.} On 2Wiki with gpt-4o-mini, an analysis of the 148 wrong answers attributed only about 19 to the DM's decisions; most of the others come from the backbone, such as nested sub-questions or wrong readings of correct passages.

\end{document}